\documentclass[letterpaper, 10 pt, conference]{ieeeconf}  

\IEEEoverridecommandlockouts                              

\usepackage[OT1]{fontenc} \usepackage{cite}
\usepackage{amsmath,amssymb,amsfonts}
\usepackage{algorithmic}
\usepackage{graphicx}
\usepackage{textcomp}
\usepackage{xcolor}
\usepackage{booktabs}
\usepackage{multirow}
\usepackage{threeparttable}
\usepackage{amssymb}
\usepackage{subfigure}
\usepackage{hyperref}
\title{\LARGE \bf
Pedestrian Trajectory Prediction via Spatial Interaction Transformer Network
}

\author{Tong Su, Yu Meng and Yan Xu
\thanks{The authors are with the School of Mechanical Engineering, 
        University of Science and Technology Beijing, China. (e-mail: 
        g20198575@xs.ustb.edu.cn; myu@ustb.edu.cn; b20160225@xs.ustb.edu.cn)
        }%
}
\begin{document}
\maketitle
\thispagestyle{empty}
\pagestyle{empty}

\begin{abstract}

As a core technology of the autonomous driving system, pedestrian trajectory prediction 
can significantly enhance the function of active vehicle safety and 
reduce road traffic injuries. In traffic scenes, when encountering with oncoming people, 
pedestrians may make sudden turns or stop immediately, which often 
leads to complicated trajectories. To predict such unpredictable trajectories, 
we can gain insights into the interaction between 
pedestrians. In this paper, we present a novel 
generative method named Spatial Interaction Transformer (SIT), 
which learns the spatio-temporal correlation of pedestrian 
trajectories through attention mechanisms. Furthermore, we introduce 
the conditional variational autoencoder (CVAE) \cite{sohn2015learning} 
framework to model the future latent motion states of pedestrians. 
In particular, the experiments based on large-scale trafﬁc dataset 
nuScenes \cite{2020nuScenes} show that SIT has 
an outstanding performance than state-of-the-art (SOTA) methods. Experimental 
evaluation on the challenging ETH \cite{5459260} and UCY \cite{lerner2007crowds} 
datasets conﬁrms the robustness of our proposed model.

\end{abstract}

\section{INTRODUCTION}

Vulnerable Roads Users(VRUs), due to their high maneuverability, may change their motion in a while. 
For the protection of pedestrians\cite{ferguson2008detection}\cite{sun2020proximity}, 
active vehicle safety systems (AVSSs) leverage the environmental 
perception and decision making technologies to minimize the effect 
of traffic accidents\cite{sun2021acclimatizing}. As a core component of AVSSs, 
pedestrian trajectory prediction module is responsible for providing warnings 
when pedestrians are close to the driving vehicles. By analyzing the 
movement patterns of other traffic agents \cite{mogelmose2015trajectory} 
and predicting their future positions, vehicles equipped with prediction 
module are able to make appropriate navigation decisions 
(e.g. avoid impending collision) \cite{park2018sequence}. 

In chaotic traffic scenes, reliable trajectory prediction is challenging 
due to pedestrians react differently according to the change of 
surrounding environment. For example, pedestrians plan their future routes 
by sensing each other's posture or subtle changes in motion \cite{yang2019top}. 
Therefore, the study of spatial interaction is essential for predicting 
future trajectories in the scenes with high pedestrian density.

\begin{figure}[htbp]
    \centering
    \includegraphics[width=\linewidth, height=4.5cm]{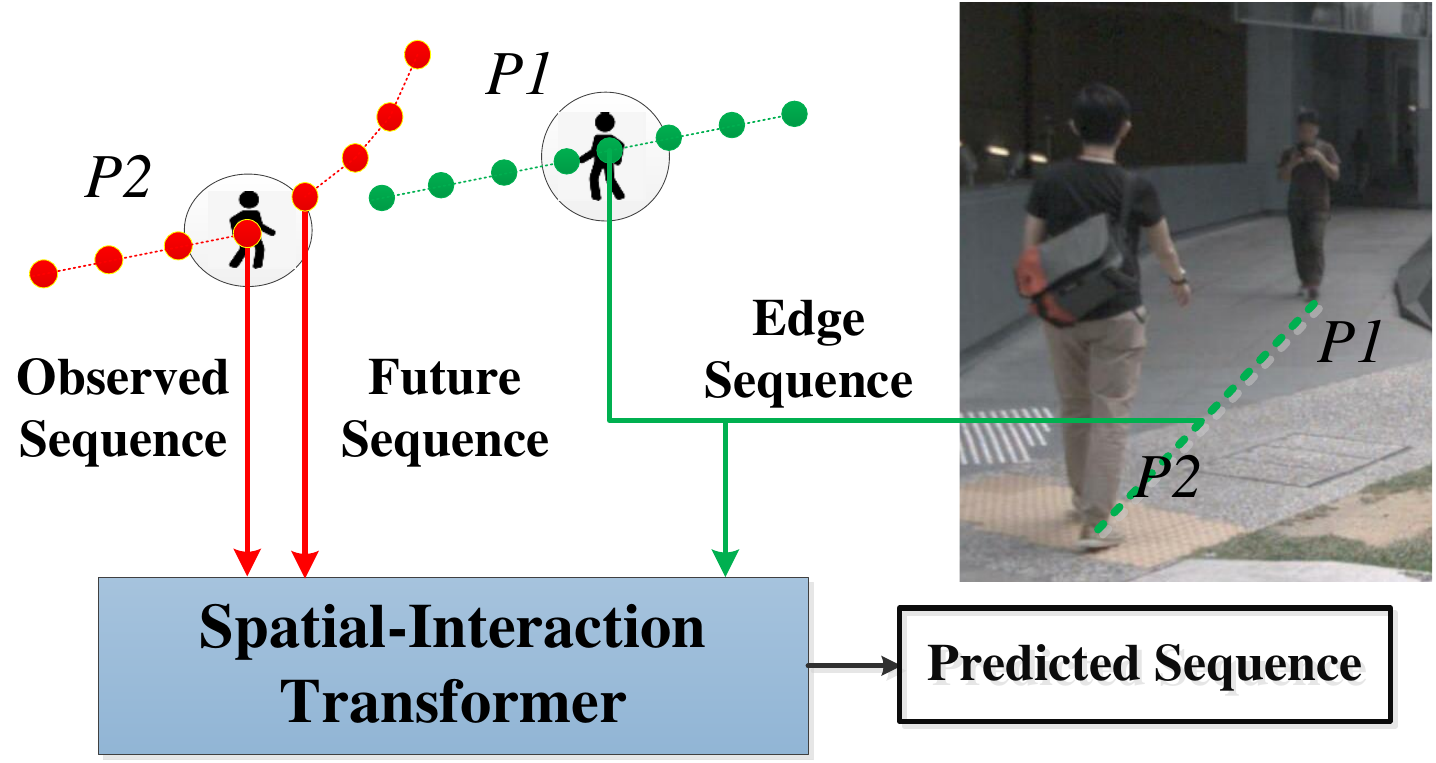}
    \caption{Scene of interest, in where we output future positions of 
    pedestrians with our proposed transformer-based methods that build a 
    social graph to model the interaction between pedestrians. Furthermore, 
    we consider the future motion states by incorporating their 
    future sequence into our model.}
    \label{fig1}
    \end{figure}

To model multi-pedestrian trajectories in the scene, we build a 
dynamic graph to capture the complex spatial interaction. Different from previous 
works \cite{alahi2016social}, transformer-based network is introduced to model 
the spatio-temporal dependencies. We believe that the powerful attention mechanisms 
are suitable for sequence modeling. Besides, we expect to aggregate 
the future latent motion states of pedestrians by their future trajectories. For example, 
in Fig. \ref{fig1}, pedestrian $\textit{P2}$ walks straight during the observation 
phase, but in the forecast phase, $\textit{P2}$ make a sudden turn to avoid 
pedestrian $\textit{P1}$. Only relying on observed trajectories is not enough 
to predict its future trajectories accurately, thus, we introduce the 
conditional variational autoencoder (CVAE) \cite{sohn2015learning} framework 
that conditions future sequence to our prediction. With this framework, 
our model is able to make full use of labels to capture the future latent motion state.

In this work, we are interested in exploring the capability of transformer-based 
network in modeling social interaction. Specifically, our network tends to 
provide feasible approaches for pedestrian trajectory prediction in heavy 
traffic environments. The main contributions are as follows: 

1): We present a novel deep generative model named 
spatial interaction transformer (SIT) that utilizes the attention mechanisms 
to dynamically model the spatial locations of the pedestrians and predict their 
future trajectories. 

2): To handle the variance of pedestrian trajectories during observation and 
prediction, we use a generative framework that follows CVAE to incorporate 
pedestrian future trajectories into social interaction. 

3): Extensive experiments are performed on two public datasets. 
Further statistical analyses show the effectiveness and robustness 
of our proposed data processing methods and edge modules.

The rest of the paper is organized as follows. In Section \ref{sec2}, 
we introduce the related work about pedestrian trajectory prediction and 
describe the transformer network for trajectory prediction. 
Section \ref{sec3} presents the detailed structure of our proposed SIT. 
The experiments and ablation study are performed in Section \ref{sec4}.
Finally, Section \ref{sec5} concludes our paper and provides 
the plans for future work.

\section{RELATED WORK \label{sec2}}
There are a large number of published studies that describe various 
pedestrian trajectory prediction methods. This 
section aims to focus on the literature relevant to our research. 
For this purpose, we introduce two aspects of related work, that 
are (a) pedestrian trajectory prediction and (b) transformer 
network for trajectory forecasting.
\subsection{Pedestrian Trajectory Prediction \label{sec2.1}}

\textbf{Traditional approaches}:$\,$Pedestrian trajectory prediction has attracted 
much attention in recent years \cite{rudenko2020human}. Forecasting methods are mainly divided 
into two categories: kinetic-based forecasting methods and data-driven 
methods \cite{DTP}. In \cite{schneider2013pedestrian}, 
Schneider et al. used the extended Kalman filter and interactive multi-model to predict 
the diverse pedestrian motion states (stationary, 
interactive, bending, starting) and analyzed their differences. 
Since Long Short-Term Memory (LSTM) is more effective in modeling long-term 
sequence than kinetic-based methods, it has become the most frequently used model in 
trajectory prediction field. In \cite{LSTMComparativeStudy}, Li et al. compared the 
effects of different learning-based methods like Gaussian Process (GP), LSTM, GP-LSTM, 
Character-based LSTM, Sequence-to-Sequence (Seq2Seq) and attention-based Seq2Seq, and 
showed that the encoder-decoder structure such as Seq2Seq has an outstanding performance 
for modeling both linear or non-linear patterns. 

\textbf{Spatial interaction}:$\,$The moving routes of pedestrians 
in traffic scenes are simply affected by surrounding agents. 
Some researchers have realized that by fusing the latent motion clues of 
surrounding pedestrians into trajectory prediction, it is possible to 
capture the dynamic changes of predicted trajectories. 
So far, social interaction has been extensively investigated 
by many works \cite{Rasouli_2020_arxiv}. Various pedestrian trajectory 
analysis techniques have been proposed, ranging from deterministic linear regression 
\cite{alahi2016social} to generative model \cite{gupta2018social}.
Social LSTM \cite{alahi2016social} encodes the historical trajectories of all 
pedestrians equally in the same scene and pools the obtained feature vectors by social pooling 
layer, which implicitly models the interaction of different pedestrians 
by learning spatial correlation. GRIP regards traffic 
objects as nodes and builds dynamic graphs to learn the movement patterns of pedestrians
\cite{li2019grip}. To predict the positions of all agents in the same scene 
at the same time, a spatio-temporal graph convolutional network is 
introduced to process the interaction of traffic subjects.

\subsection{Transformer Network for Trajectory Prediction \label{sec2.2}}
Although LSTM \cite{LSTMComparativeStudy} has been applied in a variety of 
situations, it is difficult for LSTM to enhance its computational 
speed and performance due to its sequential structure. 
Transformer \cite{vaswani2017attention} network, as the SOTA 
models for most natural language processing tasks, can rely on 
powerful attention mechanisms to avoid these shortcomings. 
By feeding past positions into the network at the same timestep, 
transformer network has superior parallel computing capabilities 
and can learn concerning information from any historical position. 
As a result, transformer network has great potential to 
achieve remarkable performance in the field of pedestrian trajectory prediction. 

Recently, Fran et al. \cite{giuliari2021transformer} applied the transformer network to 
trajectory prediction. After inferencing in the velocity space, 
they used the predicted velocity to get the coordinate position of 
the pedestrians. However, they only considered the case of modeling 
a single pedestrian. In order to cope with crowd trajectory prediction, 
we propose a new transformer-based model to simulate the interaction 
between pedestrians. 

\section{METHODOLOGY \label{sec3}}

\begin{figure*}[htbp]
    \centering
    \includegraphics[width=\textwidth]{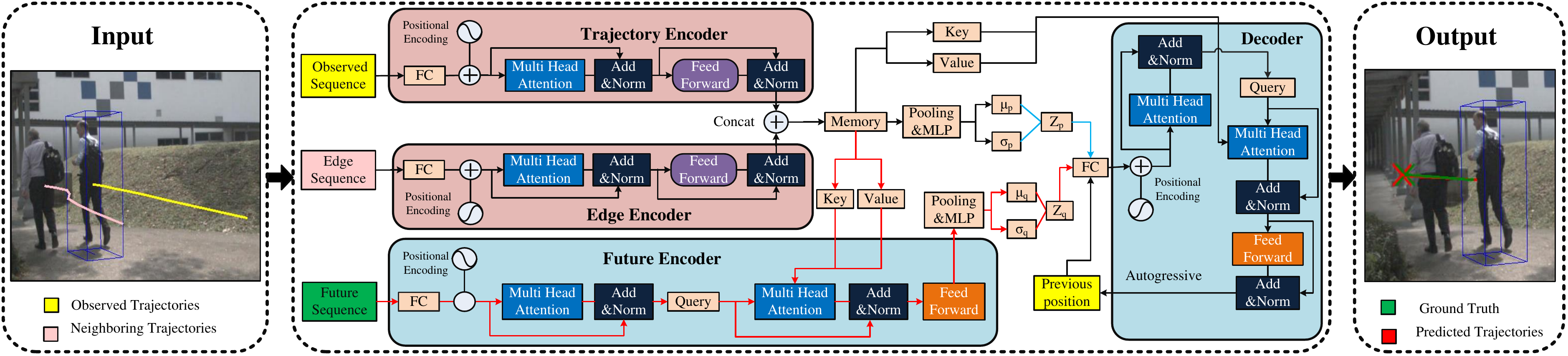}
    \caption{The details of our proposed SIT. Based on CVAE, our model 
    encodes observed sequence and their neighboring sequence to extract 
    the temporal and spatial dependencies of pedestrians respectively. Red, blue, 
    black arrows represent training only, inferencing only, and both of them, respectively.}
    \label{fig2}
    \end{figure*}

In this section, we regard the trajectory prediction problem as a sequence 
regression task. Based on the historical observation, we aim to predict 
the trajectories of pedestrians in the future timesteps. Our proposed 
probabilistic generative model SIT is shown in Fig. \ref{fig2}. 

\subsection{Input and Output of Model \label{sec3.1}} \textbf{Data processing}:$\,$In 
order to learn the distribution of trajectory samples more effectively, processing the 
trajectory data is an essential part. The most commonly adopted 
trajectory processing method is to use the mean and standard 
deviation of all pedestrian positions to normalize the data 
\cite{giuliari2021transformer}. However, in the different scenes, 
there is a large difference between the distributions of pedestrian trajectories. 
This method can not make predictions well in multiple scenarios 
\cite{ivanovic2019trajectron}. Different from it, we use each 
pedestrian's last position of observed trajectories 
as the mean and the attention radius as the standard deviation 
to normalize each sample data separately, which makes the 
input data distribution more compact and improves the accuracy of 
prediction in various scenes. 

\textbf{Input and output sequence}:$\,$ Given the 
previous $H$ timesteps, we aim to predict trajectory 
horizon of $P$ timesteps with the time interval of $\bigtriangleup t$. 
At time $t$, the observed sequence is described as 
$X_{obs}=\left \{ x^t \right \}_{t-H+1}^{t}$, future sequence 
is $Y=\left \{ y^t \right \}_{t+1}^{t+P}$ and 
predicted sequence is $\hat{Y} =\left \{ \hat{y} ^t \right \}_{t+1}^{t+P}$. 
To make full use of abundant pedestrian motion state, we 
input 6-dimensional vectors 
including the normalized positions $(x_{pos}^{t}, y_{pos}^{t})$, 
velocities $(x_{vel}^{t}, y_{vel}^{t})$, 
and accelerations $(x_{acc}^{t}, y_{acc}^{t})$. 
The predicted trajectories can generate more variability 
by adding velocity and acceleration feature to input data. 
Then, our model directly outputs the position $y^{t}$. 

\begin{align}
    x^{t} &= [x_{pos}^{t}, y_{pos}^{t}, x_{vel}^{t},y_{vel}^{t}, x_{acc}^{t},y_{acc}^{t}] \\
    y^{t} &= [x_{pos}^{t}, y_{pos}^{t}] \\
    \hat{y}^{t} &= [\hat{x}_{pos}^{t}, \hat{y}_{pos}^{t}]
\end{align}

\textbf{Positional encoding}:$\,$Unlike LSTM inputting the trajectory data step by step, 
SIT feeds into the input data simultaneously. Therefore, 
we manually add the timestamp information by using sine and cosine functions. 
Here d is each dimention of embedding.
\begin{align}
    P_{obs}     &=\begin{cases}
        sin(\frac{t}{10000^{d/D}})\quad for \quad d \quad even  \\
        cos(\frac{t}{10000^{d/D}})\quad for \quad d \quad odd 
    \end{cases}
\end{align}

\textbf{Edge sequence}:$\,$In order to predict the positions of pedestrians 
accurately, we first build a graph $G=(V,E)$ to dynamically simulate 
the interaction between pedestrians and their neighbors. Each pedestrian 
is represented as a node $\nu \in V$ and edge $e=(\nu_{i}, \nu_{j}) \in E$ exist 
when $\nu_{i}$ influences 
$\nu_{j}$. In this work, we pay attention to the neighbors within the attention 
radius like other works \cite{alahi2016social} and euclidean distance indicate 
the edge $e$. For instance, the edge $e$ is taken into account when 
$\left \| p_{i} - p_{j} \right \| \le $ attention radius 
where $p_{i}$, $p_{j} \in R^{2}$ are the 2D 
spatial position of $\nu_{i}, \nu_{j}$. In particular, the position $p_{i}$ is used to 
normalize the historical positions of neighbors $p_{j}$. This special 
technique can make all nodes in the scene have same latent states, 
which is beneficial to model the spatial interaction. 

For every single pedestrian $\nu_{i}$, we merge its past states (normalized positions, 
velocities, and accelerations) of neighbors $\nu_{j}$ when edge $e$ is present, 
which can be achieved by element-wise sum. In this way, 
we convert the variable length neighbor states to a fixed 
edge sequence $X_{edge}$ which has the same shape as the observed sequence 
$X_{obs}$.

\subsection{Spatial Interaction Transformer Network \label{sec3.2}}
The spatial interaction transformer network is mainly composed of 
attention mechanisms that assign unequal importance to neighboring 
pedestrians. Furthermore, our model follows the conditional variational 
autoencoder (CVAE) by modeling the future pedestrian trajectories 
as distributions based on their own and neighbors's past trajectories. 
We aim to learn the probabilistic model $p(Y|X_{obs}, X_{edge})$ by introducing latent 
variable $Z$. Here, $Z$ represents the latent state of pedestrian trajectories. 
We can describe the future trajectory distributions as the following equation:

\begin{align} \begin{aligned}
    p(Y|X_{obs}, X_{edge}) = & \int p(Y|X_{obs}, X_{edge}, Z) \\
    & p(Z|X_{obs}, X_{edge})dZ 
\end{aligned} \end{align}
where 
$p(Z|X_{obs}, X_{edge})$ is the gaussian prior distribution which is inferred 
by past observed sequence $X_{obs}$ and edge sequence $X_{edge}$. 
$p(Y|X_{obs}, X_{edge}, Z)$ is the conditional likelihood distribution that is 
impossible to calculate directly. So, to tackle this problem, we use the 
Kullback-Leibler divergence loss(KL loss) \cite{{sohn2015learning}} 
as one item of our loss function:

\begin{align} \begin{aligned}
    L_{kl} = KL(q(Z|Y, X_{obs}, X_{edge})\parallel p(Z|X_{obs}, X_{edge}))
\end{aligned} \end{align}
where 
$q(Z|Y, X_{obs}, X_{edge})$ is the approximate posterior distribution. The latent 
state $Z$ can be jointly inferred by the future sequence $Y$, observed 
sequence $X_{obs}$ and edge sequence $X_{edge}$. This design allows $Z$ 
to consider not only its own future trajectories but also its neighboring 
trajectories, which enables our model to generate more interactive trajectories. 
$KL$ quantifies the difference 
between two probability distributions. By minimizing the $KL$, we can 
approximate the prior and posterior distributions. During training, 
we can infer $Z$ through $q(Z|Y, X_{obs}, X_{edge})$ and during testing 
$Z$ can be inferred by $p(Z|X_{obs}, X_{edge})$. After above formulation, 
we now introduce the details of our model.

\textbf{Trajectory and Edge Encoder}:$\,$We first add the edge sequence 
and observed sequence into the timestamp information through positional encoding 
and get the embedding vector. Both of them are fed into the trajectory 
encoder and edge encoder. Specifically, we encode the observed embeddings 
and edge embeddings with Multi-head attention and feed-forward network. 
After encoding, the concatenation operation is used to get the memory vector 
$C=\left \{ c^t \right \}_{t-H+1}^{t}$ where $C$ summaries the past trajectories 
and the influence of all neighboring pedestrians. Then, a mean pooling layer is 
performed across all historical timesteps to get the past trajectory feature 
$c_{n}=mean(c_{t-H+1}, \dots, c_{t} )$. 
We use a multi layer perception (MLP) to map $c_{n}$ to the 
gaussian prior distribution $p(Z|X_{obs}, X_{edge})$ and get the gaussian parameters 
$(\mu _{p}, \sigma _{p})$. According to the Gumbel-Softmax reparameterization 
\cite{sohn2015learning}, we can sample $Z_{p}$ from the latent states:

\begin{align}
Z_{p}=\mu_{p}\epsilon + \sigma_{p}, \quad  \epsilon \sim  N(0, 1)
\end{align}

\textbf{Future Encoder}:$\,$Given the pedestrian future trajectory sequence $Y$, 
we can obtain the timestamped sequence by positional encoding. 
After encoding by Multi-head attention, this sequence is feed into 
another Multi-head attention served as queries. At the same time, the past trajectory 
memory $C$ is encoded as keys and values. The keys represent the weights for 
different timesteps and the values represent the latent state of different 
timesteps. This design allows our model to condition $X_{obs}$ through $C$, 
which is beneficial to approximate the posterior distribution effectively. 
Similar to prior distribution, a mean pooling layer is performed across 
future timesteps to extract the future feature and we use MLP to map future 
feature to the approximate posterior distribution $q(Z|Y, X_{obs}, X_{edge})$. 
Finally, we get the gaussian parameters $(\mu _{q}, \sigma _{q})$ and sample $Z_{q}$ 
from those parameters:

\begin{align}
Z_{q}=\mu _{q}\epsilon + \sigma _{q}, \quad  \epsilon \sim  N(0, 1)
\end{align}

\textbf{Future Decoder}:$\,$It is worth noting that our future decoder is 
autoregressive, which means it outputs trajectory one step at a time. 
The input sequence of decoder can be described as 
$\left \{ f^t \right \}_{t+1}^{t+P} = \left \{\hat{y}^t \oplus Z \right \}_{t+1}^{t+P}$
Here, $\hat{y}^{t+1}$ is initialized from the last position feature of 
observed sequence $X_{obs}$ and $Z$ is the sample $Z_{p}$ (training) or $Z_{q}$ (testing). 

In the decoder stage, we add timestamp information into $f^t$ through 
positional encoding and feed them into the Multi-head attention. 
After obtaining the query vector, we input it 
into another Multi-head attention along with past trajectory memory served as 
keys and values. Then feed forward network is applied to output next timestep 
trajectory. Our future decoder allows the model to inference future trajectories 
while considering the effect of current neighbors. To approximate the conditional 
likelihood distribution $p(Y|X_{obs}, X_{edge}, Z)$ according to 
$q(Z|Y, X_{obs}, X_{edge})$, we minimize the mean squred error between 
predicted trajectories and future trajectory labels. 
Our loss function is written as:
\begin{align}
    L & = \min \parallel Y-\hat{Y} \parallel + L_{kl}
    \end{align}

\subsection{Training Details \label{sec3.3}}

We use batch size 100 on the training set and testing set. The latent states 
$|Z|=32$. A 3-layer encoder and a 3-layer decoder are applied to our network. 
For data augmentation, we rotate the observed trajectories by an angle that varies from 
$0^\circ$ to $360^\circ$ with an interval of $15^\circ$. 

During training, we maintain the same weight initialization for all layers of SIT. 
The same as \cite{giuliari2021transformer}, we use 8 heads for Multi-head attention 
and D = 256. Following previous work\cite{alahi2016social}, attention radius 
is set to 10 meters. We also use Adam as our optimizer with a decaying learning rate.
Then the SIT is trained for 100 epochs.

\section{EXPERIMENTS \label{sec4}}

In this section, we evaluate our proposed method on two public 
datasets. Besides, we perform the 
ablation study to quantitatively describe the effects of different 
components. It is worth noting that the experiments include varying 
prediction horizons.

\subsection{Datasets \label{sec4.1}}

We mainly use two public datasets to verify the performance 
of our proposed method. Both of them are in the world coordinate 
system. The first is the large-scale traffic dataset nuScenes 
for autonomous driving, which is annotated at 2 frames per second 
($\bigtriangleup t$ = 0.5). 
We extract scenes that contain pedestrians and get a 
training set of 632 scenes and a test set of 133 scenes. 
Each scene is 20s long. 
The total training sample contains 75767 sequences, 
and the test sample contains 12876 sequences. 
For the nuScenes dataset, we predict the trajectories 
of future 3s ($H$=6) based on the 
observed trajectories of past 4s ($P$=8). 
The second is the widely used benchmark datasets 
ETH (ETH and HOTEL) and UCY (UNIV, ZARA1, and ZARA2) 
in the field of pedestrian trajectory prediction. 
The datasets contain 5 different scenarios, each of which 
contains complex pedestrian interaction behaviors and is 
annotated at 2.5 Hz ($\bigtriangleup t$ = 0.4). 
In order to maintain a fair comparison, we predict the 
trajectory horizon of 8 timesteps (3.2s) based on the observed length 
of 12 timesteps (4.8s) same as most papers \cite{gupta2018social}.

\subsection{Evaluation Metrics \label{sec4.2}}

The same as prior papers \cite{ma2019trafficpredict, bhattacharyya2018long}, 
we use the following metrics to evaluate our methods:

MAD (Mean Average Displacement, equivalently Average 
Displacement Error ADE):$\,$The average 
value of the Euclidean distance error between the 
predicted trajectories and the ground truth in the future 
time horizon T.

FAD (Final Average Displacement, equivalently Final 
Displacement Error FDE):$\,$The average value 
of the Euclidean distance error between the predicted
trajectories and the ground truth at the last time 
step.
\subsection{NuScenes Dataset \label{sec4.3}}

\textbf{Experiment results}: To evaluate the effectiveness of our model, 
We compared the deterministic version of our proposed SIT with other 
baseline methods:

\begin{enumerate}
\item Constant Velocity(CV): This model assumes that 
pedestrians move at a constant velocity to 
linearly reason about the future trajectories.
\item LSTM: An LSTM encoder-decoder 
predicts future locations by using the past 
trajectories of pedestrians.
\item LSTM+Attention: Based on the LSTM 
encoder-decoder, an attention layer is added 
to the encoder. The attention mechanisms can automatically search 
for parts of the observed trajectories which are closely related to the 
the predicted sequence.
\item Transformer \cite{giuliari2021transformer}: 
We keep the original model completely but use the nuScenes dataset for 
pedestrian trajectory prediction task.
\end{enumerate}

Following the above methods, we separately conduct the 
experiments on the nuScenes dataset to compare the performance
with prediction horizon varying from 1s to 3s.
As shown in table \ref{table1} and Fig. \ref{fig3}, 
the prediction error MAD and FAD increase
with the growth of time horizon. In addition, our model has 
lowest prediction errors over baseline methods (outperforming existing 
approach \cite{giuliari2021transformer} by 53\% on MAD in the case of predicting 3s). 

\begin{center}
\begin{table}[htbp]
      \centering
      \caption{Comparison among above mentioned methods(lower is better)}
      \begin{threeparttable}
      \setlength{\linewidth}{3mm}{
            \begin{tabular}{c|ccc|ccc}
            \toprule
            \multirow{2}[3]{*}{Methods} & \multicolumn{3}{c|}{MAD(meters)} & \multicolumn{3}{c}{FAD(meters)} \\
            \cmidrule{2-7} & 1s    & 2s    & 3s  & 1s    & 2s    & 3s \\
            \midrule
            Const.Velocity & 0.16  & 0.27  & 0.39  & 0.21  & 0.44  & 0.69  \\
            LSTM  & 0.14  & 0.24  & 0.35  & 0.18  & 0.39  & 0.63  \\
            LSTM+Attention & 0.14  & 0.24  & 0.34  & 0.18  & 0.39  & 0.62  \\
            Transformer\cite{giuliari2021transformer} & 0.11  & 0.20  & 0.30  & 0.15  & 0.34  & 0.57  \\
            \textbf{SIT(ours)}   & \textbf{0.03}  & \textbf{0.08}  & \textbf{0.16}  & \textbf{0.03}  & \textbf{0.17}  & \textbf{0.36}  \\
            \bottomrule
            \end{tabular}}%
      \label{table1}%
      \end{threeparttable}
\end{table}%
    \end{center} 

\begin{figure}[htbp]
    \centering
    \includegraphics[width=\linewidth, height=4cm]{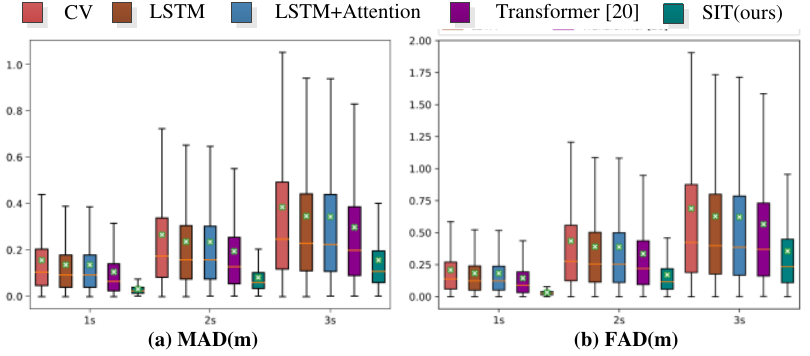}
    \caption{The quantitative MAD and FAD results of all methods 
    on the nuScenes test set when predicting 1-3s. We sample all test 
    trajectories and use the boxplots to describe the distributions of their 
    mean errors. "x" markers indicate the MAD or FAD value.}
    \label{fig3}
\end{figure}

\begin{figure}[htbp]
    \centering
    \includegraphics[width=0.95\linewidth, height=5.5cm]{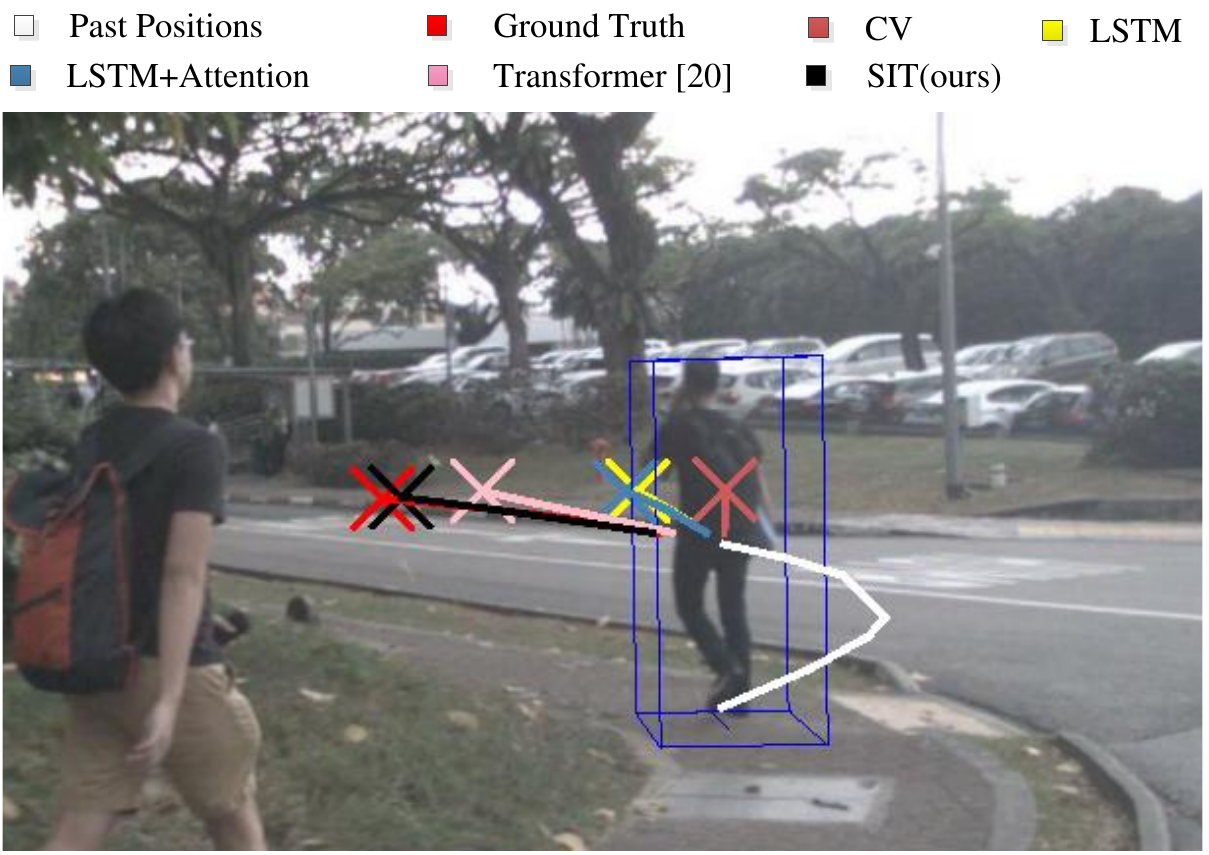}
    \caption{Predicted trajectory examples for different baseline methods. 
    The blue box represents the target pedestrian. The white line is 
    the past positions and the red line is the ground truth of future positions. 
    The other colors represent the respective trajectories predicted by different methods. 
    We can see that SIT is closer to the ground truth.}
    \label{fig4}
    \end{figure}

Then, we project the predicted trajectories to the image plane to 
visualize deterministic predicted examples of above mentioned prediction methods. 
As shown in Fig. \ref{fig4}, our proposed SIT performs particularly 
well in situations where a pedestrian may begin to turn suddenly. Our 
CVAE method plays a major role in modeling the nonlinear dynamics.

\textbf{Ablation study}: A comprehensive ablation study is performed in table 
\ref{table2}. We first make a qualitative analysis on the 
basis of the original transformer 
(the first row of table \ref{table2}) \cite{giuliari2021transformer}. 
To fairly compare the impact of different components, we adopt the data 
processing method as described in \ref{sec3.1}. The performance of original 
transformer data processing method can be seen in the fourth row of \ref{table1}. 
There is a slight improvement when adding to the edge encoding, which 
is crucial for modeling the spatial interaction. 
As can be seen in the third row, the CVAE framework yields 
a drastic reduction whether in MAD or FAD.

\begin{center}
\begin{table}[htbp]
    \centering
    \caption{ABLATION STUDY}
    \begin{threeparttable}
    \setlength{\linewidth}{3mm}{
        \begin{tabular}{cc|ccc|ccc}
        \toprule
        \multicolumn{2}{c|}{Components} & \multicolumn{3}{c|}{MAD} & \multicolumn{3}{c}{FAD} \\
        \midrule
        E$^{\mathrm{a}}$     & C$^{\mathrm{b}}$   & \multicolumn{1}{c}{1s} & \multicolumn{1}{c}{2s} & \multicolumn{1}{c|}{3s} & \multicolumn{1}{c}{1s} & \multicolumn{1}{c}{2s} & \multicolumn{1}{c}{3s} \\
        \midrule
        - & - & 0.07 & 0.13 & 0.22 & 0.09 & 0.24 & 0.46  \\
        \checkmark & - & 0.06 & 0.11 & 0.19 & 0.07 & 0.20  & 0.39  \\
        \checkmark & \checkmark & \textbf{0.03}  & \textbf{0.08} & \textbf{0.16} & \textbf{0.03} & \textbf{0.17} & \textbf{0.36}  \\
        \bottomrule
        \end{tabular}}%
    \label{table2}%
    \begin{tablenotes}
        \footnotesize
        \item a. Edge encoding;  b. CVAE 
        \end{tablenotes}
    \end{threeparttable}
\end{table}%

    \end{center} 

\subsection{ETH and UCY Datasets \label{sec4.4}}
To verify the robustness of our proposed method, 
we perform corresponding verifications on the benchmark 
datasets ETH and UCY in the prediction field. 
The leave-one-out evaluation strategy is 
generally adopted by most works. Specifically, we use 4 datasets 
of ETH and UCY to train the model and the rest to test. 

As presented in table \ref{table3}, the deterministic model output 
one single trajectory and except ETH dataset our model has 
achieved outstanding performance among all SOTA methods. The 
performance of stochastic model is summaried in table \ref{table4}. 
Here we sample 20 times and report the best sample. We can observe that our 
SIT significantly outperforms the baselines whether in MAD or FAD. 
One interesting finding is that our model significantly outperforms on 
UCY and ZARA, where crowd density is relatively higher. This can be explained by 
that our model is suitable to model the human-huamn interaction.

\begin{center}
\begin{table}[htbp]
  \centering
  \caption{deterministic model}
  \begin{threeparttable}
  \setlength{\tabcolsep}{0.5mm}{
    \begin{tabular}{crccrccrccrcc}
    \toprule
    \multirow{2}[1]{*}{Datasets}  &  & \multicolumn{2}{c}{S-LSTM\cite{alahi2016social}} &       & \multicolumn{2}{c}{TF$^{\mathrm{c}}$ \cite{giuliari2021transformer}} &       & \multicolumn{2}{c}{STAR-D\cite{yu2020spatio}} &       & \multicolumn{2}{c}{\textbf{SIT(ours)}} \\
    \cmidrule{3-4}\cmidrule{6-7}\cmidrule{9-10}\cmidrule{12-13} &       & M$^{\mathrm{a}}$   & F$^{\mathrm{b}}$   &       & M   & F   &       & M   & F   &          & \textbf{M}   & \textbf{F} \\
    \midrule
    ETH   &  & 1.09  & 2.35  &  & 1.03  & 2.10  & & \textbf{0.56}  & \textbf{1.11}  &  & 0.59  & 1.28 \\
    HOTEL &  & 0.79  & 0.76  &  & 0.36  & 0.71  & & 0.26  & 0.50  &  & \textbf{0.22}  & \textbf{0.45} \\
    UCY   &  & 0.67  & 1.40  &  & 0.53  & 1.32  & & 0.52  & 1.15  &  & \textbf{0.40}  & \textbf{0.98} \\
    ZARA1 &  & 0.47  & 1.00  &  & 0.44  & 1.00  & & 0.41  & 0.90  &  & \textbf{0.30}  & \textbf{0.75} \\
    ZARA2 &  & 0.56  & 1.17  &  & 0.34  & 0.76  & & 0.31  & 0.71  &  & \textbf{0.23}  & \textbf{0.59} \\
    \midrule
    avg$^{\mathrm{d}}$   &  & 0.72  & 1.54  &       & 0.54   & 1.17  &       & 0.41  & 0.87  &        & \textbf{0.35}  & \textbf{0.81} \\
    \bottomrule
    \end{tabular}}%
  \label{table3}%
  \begin{tablenotes}
  \footnotesize
  \item a. MAD;  b. FAD;  c. Transformer
  \item d.The average value of 5 datasets
  \end{tablenotes}
\end{threeparttable}
\end{table}%

    \end{center}

\begin{center}
\begin{table}[htbp]
  \centering
  \caption{stochastic model(best of 20 samples)}
  \begin{threeparttable}
  \setlength{\tabcolsep}{0.5mm}{
    \begin{tabular}{crccrccrccrcc}
    \toprule
    \multirow{2}[1]{*}{Datasets}  &  & \multicolumn{2}{c}{S-GAN\cite{gupta2018social}} &       & \multicolumn{2}{c}{TF \cite{giuliari2021transformer}} &       & \multicolumn{2}{c}{STAR\cite{yu2020spatio}} &        & \multicolumn{2}{c}{\textbf{SIT(ours)}} \\
    \cmidrule{3-4}\cmidrule{6-7}\cmidrule{9-10}\cmidrule{12-13} &       & M   & F   &       & M   & F   &       & M   & F   &         & \textbf{M}   & \textbf{F} \\
    \midrule
    ETH   &  & 0.81  & 1.52  &  & 0.61  & 1.12  & & \textbf{0.36}  & \textbf{0.65}  &   & 0.38  & 0.88 \\  
    HOTEL &  & 0.72  & 1.61  &  & 0.18  & 0.30  & & 0.17  & 0.36  &   & \textbf{0.11}  & \textbf{0.21} \\
    UCY   &  & 0.60  & 1.26  &  & 0.35  & 1.65  & & 0.31  & 0.62  &   & \textbf{0.20}  & \textbf{0.46} \\
    ZARA1 &  & 0.34  & 0.69  &  & 0.22  & 0.38  & & 0.26  & 0.55  &   & \textbf{0.16}  & \textbf{0.37} \\
    ZARA2 &  & 0.42  & 0.84  &  & 0.17  & 0.32  & & 0.22  & 0.46  &   & \textbf{0.12}  & \textbf{0.27} \\
    \midrule
    avg   &  & 0.58  & 1.18  &     & 0.31  & 0.55  &       & 0.26  & 0.53  &       & \textbf{0.19}  & \textbf{0.44} \\
    \bottomrule
    \end{tabular}}%
  \label{table4}%
\end{threeparttable}
\end{table}%

    \end{center}
\section{DISCUSSION AND FUTURE WORK \label{sec5}}

This research aims to provide a novel method for 
pedestrian trajectory prediction task. Specifically, 
we propose deep generative model SIT which is based on attention mechanisms. 
In contrast to LSTM’s sequential structure, our model learns 
the spatio-temporal correlation at a deeper level thanks to the 
self-attention network. We also merge the ground truth of future 
sequence into our trajectory prediction model. With CVAE, our model 
can make full use of future sequence. The effectiveness of our proposed 
modules is proved by our ablation study. On the other hand, we explore the ability 
of the transformer-based model to encode the spatial interaction between pedestrians. 
The results suggest that our method can achieve 
a significant improvement than SOTA methods \cite{giuliari2021transformer}.

Only adopting the pedestrian trajectories to make predictions, 
existing methods might fail in some complex scenes. In the development of 
the paper, we focus on fusing extra information like 
vehicles \cite{xue2019crossing} and scenes \cite{makansi2020multimodal}
through transformer network. Moreover, future work 
will also pay attention to the variable length history trajectories 
because missing values \cite{yu2020spatio} often exist in real world. 
We expect that the transformer network can solve this 
problem without padding or interpolation. 

\addtolength{\textheight}{-12cm}   




\section*{ACKNOWLEDGMENT}

This work was supported by the National key Research and Development Project of China 
(grant no. 2018YFE0192900). This work was presented at the 2nd Workshop on Naturalistic 
Road User Data and its Applications for Automated Driving (WS11), IV2021.


\bibliographystyle{IEEEtran.bst}
\bibliography{IVWS_2021_TrajPred.bib}

\end{document}